\documentclass[11pt,a4paper]{article}
\usepackage[hyperref]{acl2019}
\usepackage{times}
\usepackage{latexsym}
\usepackage{url}
\usepackage{paralist}
\usepackage{multirow}
\usepackage{verbatim}
\usepackage{amsfonts}
\usepackage{amssymb}
\usepackage{amsmath}
\usepackage{algorithm}
\usepackage{algpseudocode}
\usepackage{graphics}
\usepackage{epsfig}

\aclfinalcopy % Uncomment this line for the final submission
\DeclareMathOperator*{\argmax}{arg\,max}

\title{Dual Supervised Learning for \\ Natural Language Understanding and Generation}

\author{Shang-Yu Su \quad Chao-Wei Huang \quad Yun-Nung Chen \\
Department of Computer Science and Information Engineering\\
National Taiwan University\\
\texttt{ \{f05921117,r07922069\}@ntu.edu.tw\quad y.v.chen@ieee.org }}

\date{}

\begin{document}
\maketitle
\begin{abstract}
Natural language understanding (NLU) and natural language generation (NLG) are both critical research topics in the NLP and dialogue fields.
Natural language understanding is to extract the core semantic meaning from the given utterances, while natural language generation is opposite, of which the goal is to construct corresponding sentences based on the given semantics. 
However, such dual relationship has not been investigated in literature.
This paper proposes a novel learning framework for natural language understanding and generation on top of dual supervised learning, providing a way to exploit the duality.
The preliminary experiments show that the proposed approach boosts the performance for both tasks, demonstrating the effectiveness of the dual relationship.\footnote{\url{https://github.com/MiuLab/DuaLUG}}
\end{abstract}

\section{Introduction}
Spoken dialogue systems that can help users solve complex tasks such as booking a movie ticket have become an emerging research topic in artificial intelligence and natural language processing areas. 
With a well-designed dialogue system as an intelligent personal assistant, people can accomplish certain tasks more easily via natural language interactions. 
% Today, there are several virtual intelligent assistants on the market, such as Apple's Siri, Google's Home, Microsoft's Cortana, and Amazon's Echo. 
The recent advance of deep learning has inspired many applications of neural dialogue systems~\cite{wen2017network,bordes2017learning,dhingra2017towards,li2017end}.
A typical dialogue system pipeline can be divided into several parts: 1) a speech recognizer that transcribes a user's speech input into texts, 2) a natural language understanding module (NLU) that classifies the domain and associated intents and fills slots to form a semantic frame ~\cite{chi2017speaker,chen2017dynamic,zhang2018addressee, su2018time, su2019dynamically}, 3) 
a dialogue state tracker (DST) that predicts the current dialogue state in the multi-turn conversations, 4) a dialogue policy that determines the system action for the next step given the current state~\cite{peng2018deep, su2018discriminative}, and 5) a natural language generator (NLG)
that outputs a response given the action semantic frame~\cite{wen2015semantically, su2018natural, su2018investigating}.

\begin{figure}[t]
\centering 
\includegraphics[width=\linewidth]{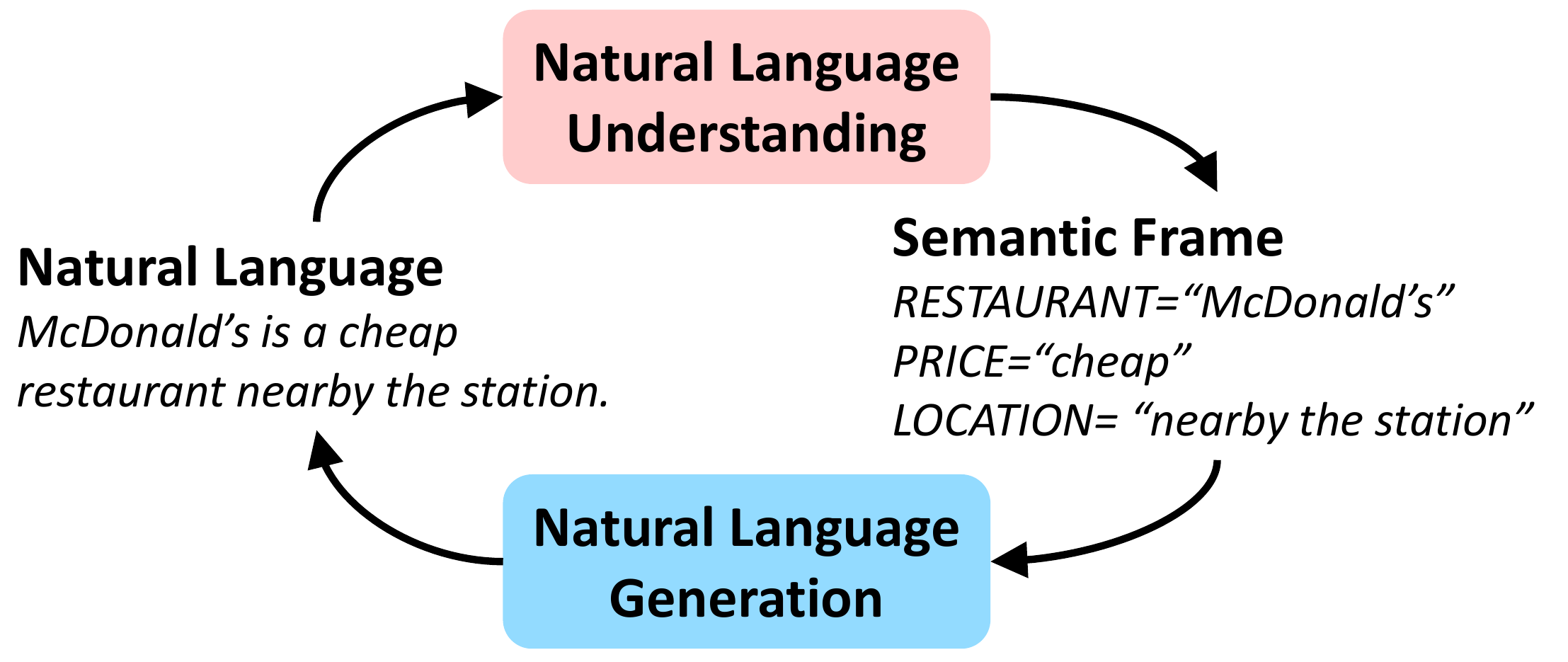}
\vspace{-3mm}
\caption{NLU and NLG emerge as a dual form.} 
\label{fig:framework} 
\vspace{-3mm}
\end{figure}

Many artificial intelligence tasks come with a \emph{dual} form; that is, we could directly swap the input and the target of a task to formulate another task.
Machine translation is a classic example \cite{wu2016google}; for example, translating from English to Chinese has a dual task of translating from Chinese to English; automatic speech recognition (ASR) and text-to-speech (TTS) also have structural duality \cite{tjandra2017listening}. 
Previous work first exploited the duality of the task pairs and proposed supervised \cite{xia2017dual} and unsupervised (reinforcement learning) \cite{he2016dual} training schemes. 
The recent studies magnified the importance of the duality by boosting the performance of both tasks with the exploitation of the duality.

NLU is to extract core semantic concepts from the given utterances, while the goal of NLG is to construct corresponding sentences based on given semantics.
In other words, understanding and generating sentences are a dual problem pair shown in Figure~\ref{fig:framework}. 
In this paper, we introduce a novel training framework for NLU and NLG based on \emph{dual supervised learning} \cite{xia2017dual}, which is the first attempt at exploiting the duality of NLU and NLG.
The experiments show that the proposed approach improves the performance for both tasks.

\section{Proposed Framework}
This section first describes the problem formulation, and then introduces the core training algorithm along with the proposed methods of estimating data distribution.

\begin{comment}
\subsection{Problem Formulation}
Suppose we have two spaces: the semantics space $\mathcal{X}$ and the space $\mathcal{Y}$ of natural language.
The goal of language generation is, given semantics, to generate corresponding utterance. In other words, the task is to learn a mapping function $f: \mathcal{X} \to \mathcal{Y}$ to transform semantic representation into natural language.
On the other hand, language understanding is to capture the core meaning of utterances, the task is to find a function $g: \mathcal{Y} \to \mathcal{X}$ to predict semantic representation from given natural language. 

Given $n$ data pairs $\{(x_i, y_i)\}^n_{i=1}$ $i.i.d.$ sampled from the joint space $\mathcal{X} \times \mathcal{Y}$.
A typical strategy of these optimization problem is based on maximum likelihood estimation (MLE) of the parameterized conditional distribution by the learnable parameters $\theta_{x \to y}$ and $\theta_{y \to x}$:

\begin{align*}
f(x;\theta_{x \to y}) = \argmax_{y' \in \mathcal{Y}} P(y' \mid x ; \theta_{x \to y} ), \\
g(y;\theta_{y \to x}) = \argmax_{x' \in \mathcal{X}} P(x' \mid y ; \theta_{y \to x} ). \\
\end{align*}
\end{comment}

Assuming that we have two spaces, the semantics space $\mathcal{X}$ and the natural language space $\mathcal{Y}$,
given $n$ data pairs $\{(x_i, y_i)\}^n_{i=1}$, the goal of NLG is to generate corresponding utterances based on given semantics.
In other words, the task is to learn a mapping function $f(x;\theta_{x \to y})$ to transform semantic representations into natural language.
On the other hand, NLU is to capture the core meaning of utterances, finding a function $g(y;\theta_{y \to x})$ to predict semantic representations given natural language. 
% Given $n$ data pairs $\{(x_i, y_i)\}^n_{i=1}$ $i.i.d.$ sampled from the joint space $\mathcal{X} \times \mathcal{Y}$.
A typical strategy of these optimization problems is based on maximum likelihood estimation (MLE) of the parameterized conditional distribution by the learnable parameters $\theta_{x \to y}$ and $\theta_{y \to x}$.

\subsection{Dual Supervised Learning}
% \subsubsection{Masked Autoencoder for Estimating Marginal Distribution}
% \subsubsection{Distribution Estimation as Autoregression}
Considering the duality between two tasks in the dual problems, it is intuitive to bridge the bidirectional relationship from a probabilistic perspective. 
If the models of two tasks are optimal, we have \textit{probabilistic duality}:
\begin{align*}
% \mathbb{E}_{P_D(x)}[\mathrm{log}p_{\theta}(x)] = 
% \mathbb{E}_{P_D(x)}[\mathrm{log}\mathbb{E}_{P(z)}[p_{\theta}(x)]] \\ 
P(x)P(y \mid x ; \theta_{x \to y}) &= P(y)P(x \mid y ; \theta_{y \to x} ) \\
&= P(x,y) \ \forall x,y, 
\end{align*}
where $P(x)$ and $P(y)$ are marginal distributions of data.
The condition reflects parallel, bidirectional relationship between two tasks in the dual problem.
Although standard supervised learning with respect to a given loss function is a straight-forward approach to address MLE, it does not consider the relationship between two tasks.

\citet{xia2017dual} exploited the duality of the dual problems to introduce a new learning scheme, which explicitly imposed the empirical probability duality on the objective function. 
The training strategy is based on the standard supervised learning and incorporates the probability duality constraint, so-called \textit{dual supervised learning}.
Therefore the training objective is extended to a multi-objective optimization problem:
\begin{align*}
\begin{cases}
\min_{\theta_{x \to y}} (\mathbb{E}[l_{1}(f(x;\theta_{x \to y}),y)]), \\
\min_{\theta_{y \to x}} (\mathbb{E}[l_{2}(g(y;\theta_{y \to x}),x)]), \\
\text{s.t.} \ P(x)P(y \mid x ; \theta_{x \to y}) = P(y)P(x \mid y ; \theta_{y \to x}),
\end{cases}
\end{align*}
where $l_{1,2}$ are the given loss functions.
Such constraint optimization problem could be solved by introducing Lagrange multiplier to incorporate the constraint:
\begin{align*}
% f(x;\theta_{x \to y}) = \argmax_{y' \in \mathcal{Y}} P(y' \mid x ; \theta_{x \to y} ), \\
% g(y;\theta_{y \to x}) = \argmax_{x' \in \mathcal{X}} P(x' \mid y ; \theta_{y \to x} ). \\
% \min_{\theta_{x \to y}} ((1/n) \sum_{i=1}^n Loss(f(x;\theta_{x \to y}),y))
\begin{cases}
\min_{\theta_{x \to y}} (\mathbb{E}[l_{1}(f(x;\theta_{x \to y}),y)] + \lambda_{x \to y} l_{duality}), \\
\min_{\theta_{y \to x}} (\mathbb{E}[l_{1}(g(y;\theta_{y \to x}),x)] + \lambda_{y \to x} l_{duality}), \\
\end{cases}
\end{align*}
where $\lambda_{x \to y}$ and $\lambda_{y \to x}$ are the Lagrange parameters and the constraint is formulated as follows:
\begin{align*}
l_{duality} &= (\mathrm{log}\hat{P}(x) + \mathrm{log}P(y \mid x ; \theta_{x \to y}) \\ 
&- \mathrm{log}\hat{P}(y) - \mathrm{log}P(x \mid y ; \theta_{y \to x} ))^2.
\end{align*}

Now the entire objective could be viewed as the standard supervised learning with an additional regularization term considering the duality between tasks.
Therefore, the learning scheme is to learn the models by minimizing the weighted combination of an original loss term and a regularization term.
Note that the true marginal distribution of data $P(x)$ and $P(y)$ are often intractable, so here we replace them with the approximated empirical marginal distribution $\hat{P}(x)$ and $\hat{P}(y)$.

\subsection{Distribution Estimation as Autoregression}
With the above formulation, the current problem is how to estimate the empirical marginal distribution $\hat{P(\cdot)}$.
To accurately estimate data distribution, the data properties should be considered, because different data types have different structural natures.
For example, natural language has sequential structures and temporal dependencies, while other types of data may not.
Therefore, we design a specific method of estimating distribution for each data type based on the expert knowledge.

From the probabilistic perspective, we can decompose any data distribution $p(x)$ into the product of its nested conditional probability,
\begin{align}
p(x) = \prod_{d}^{D} p(x_{d} \mid x_{1}, ... , x_{d-1}),
\label{eq:auto}
\end{align}
where $x$ could be any data type and $d$ is the index of a variable unit. 
% describe autoregression
\subsubsection{Language Modeling}
Natural language has an intrinsic sequential nature; therefore it is intuitive to leverage the autoregressive property to learn a language model.
In this work, we learn the language model based on recurrent neural networks \cite{mikolov2010recurrent, sundermeyer2012lstm} by the cross entropy objective in an unsupervised manner.
\begin{align}
 p(y) = \prod_{i}^{L} p(y_{i} \mid y_{1}, ... , y_{i-1}; \theta_{y}),
 \label{eq:lm}
 \end{align}
 where $y_{(\cdot)}$ are words in the sentence $y$, and $L$ is the sentence length.
% here could expand more

\subsubsection{Masked Autoencoder}
The semantic representation $x$ in our work is discrete semantic frames containing specific slots and corresponding values.
Each semantic frame contains the core concept of a certain sentence, for example, the slot-value pairs ``\texttt{name[Bibimbap House], food[English], priceRange[moderate], area [riverside], near[Clare Hall]}''  corresponds to the target sentence ``\emph{Bibimbap House is a moderately priced restaurant who's main cuisine is English food. You will find this local gem near Clare Hall in the Riverside area.}''. 
Even though the product rule in (\ref{eq:auto}) enables us to decompose any probability distribution into a product of a sequence of conditional probability, how we decompose the distribution reflects a specific physical meaning.
For example, language modeling outputs the probability distribution over vocabulary space of $i$-th word $y_i$ by only taking the preceding word sequence $y_{<i}$. 
Natural language has the intrinsic sequential structure and temporal dependency, so modeling the joint distribution of words in a sequence by such autoregressive property is logically reasonable. 
However, slot-value pairs in semantic frames do not have a single directional relationship between them, while they parallel describe the same sentence, so treating a semantic frame as a sequence of slot-value pairs is not suitable.
Furthermore, slot-value pairs are not independent, because the pairs in a semantic frame correspond to the same individual utterance.
For example, French food would probably cost more.
% Therefore, the correlation hinders from obtaining the joint distribution by product.
Therefore, the correlation should be taken into account when estimating the joint distribution.

\begin{figure}[t!]
\centering 
\includegraphics[width=.85\linewidth]{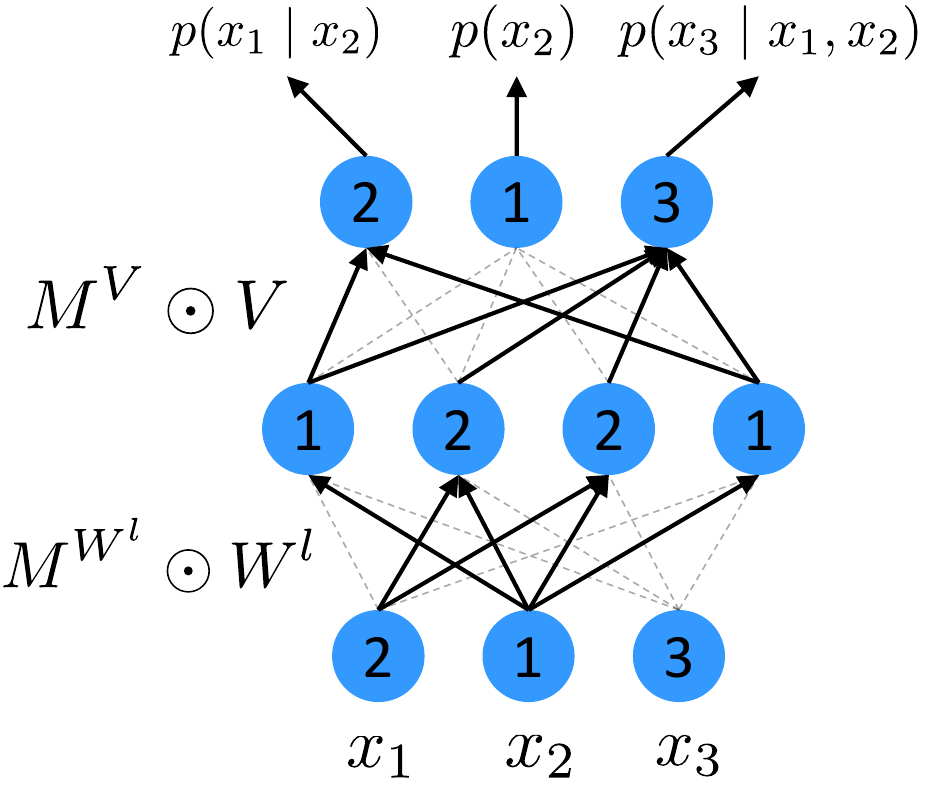}
\vspace{-1mm}
\caption{The illustration of the masked autoencoder for distribution estimation (MADE).} 
\label{fig:ste} 
\vspace{-2mm}
\end{figure}

Considering the above issues, to model the joint distribution of flat semantic frames, various dependencies between slot-value semantics should be leveraged.
In this work, we propose to utilize a masked autoencoder for  distribution estimation (MADE) \cite{germain2015made}.
By zeroing certain connections, we could enforce the variable unit $x_{d}$ to only depend on any specific set of variables, not necessary on $x_{<d}$; eventually we could still have the marginal distribution by the product rule:
\begin{equation}
p(x) = \prod_{d}^{D} p(x_{d} \mid S_{d} ),
\end{equation}
where $S_{d}$ is a specific set of variable units.

\begin{comment}
In practical, we elementwise-multiply every weight matrix by a binary mask matrix $M$ to interrupt some connections.
The idea is simple, to impose the autoregressive property we first assign each hidden unit $k$ a integer number $m(k)$ ranging from 1 to the dimension of data $D-1$ inclusively.  
For the weight matrices of hidden layer $W^{l}$, we build binary mask matrices as follows:
\begin{align*}
M^{W^l}_{} = 
\begin{cases}
1 & \text{if } m^{l}(k') \geq m^{l-1}(k), \\
0 & \text{otherwise }, 
\end{cases}
\end{align*}
where $l$ indicates the index of the hidden layer.
For the input and output layer, we assign each unit a number ranging from 1 to $D$ exclusively, then we enforce build the mask for the output matrix $V$ as
\begin{align*}
M^{V}_{} = 
\begin{cases}
1 & \text{if } m^{L}(d) > m^{L-1}(k), \\
0 & \text{otherwise }, 
\end{cases}
\end{align*}
where $L$ indicates the output layer. With the constructed mask matrices, the masked autoencoder is shown to be able to estimate the joint distribution as autoregression.
Because there is no explicit rule specifying the exact dependencies between the slot-value pairs in our data, we consider various dependencies by ensemble of multiple decomposition, that is, to sample different sets $S_{d}$. 
\end{comment}

In practice, we elementwise-multiply each weight matrix by a binary mask matrix $M$ to interrupt some connections, as illustrated in Figure~\ref{fig:ste}.
To impose the autoregressive property, we first assign each hidden unit $k$ an integer $m(k)$ ranging from 1 to the dimension of data $D-1$ inclusively; 
for the input and output layers, we assign each unit a number ranging from 1 to $D$ exclusively.
Then binary mask matrices can be built as follows:
\begin{align*}
M^{}_{} = 
\begin{cases}
1 & \text{if } m^{l}(k') \geq m^{l-1}(k), \\
1 & \text{if } m^{L}(d) > m^{L-1}(k), \\
0 & \text{otherwise.} 
\end{cases}
\end{align*}
Here $l$ indicates the index of the hidden layer, and $L$ indicates the one of the output layer.
% For the input and output layer, we assign each unit a number ranging from 1 to $D$ exclusively, then we enforce build the mask for the output matrix $V$ as follows:
% \begin{align*}
% M^{V}_{} = 
% \begin{cases}
% 1 & \text{if } m^{L}(d) > m^{L-1}(k), \\
% 0 & \text{otherwise }, \\
% \end{cases}
% \end{align*}
% where $L$ indicates the output layer. 
With the constructed mask matrices, the masked autoencoder is shown to be able to estimate the joint distribution as autoregression.
Because there is no explicit rule specifying the exact dependencies between slot-value pairs in our data, we consider various dependencies by ensemble of multiple decomposition, that is, to sample different sets $S_{d}$.

% EXP RESULT TABLE
\begin{table*}
\centering
% \small
\begin{tabular}{ | c| l | c | c c c c| }
    \hline
    \multicolumn{2}{|c|}{\multirow{2}{*}{\bf Learning Scheme}} & \bf NLU & \multicolumn{4}{c|}{\bf NLG}\\
     \multicolumn{2}{|c|}{} & \bf \small F1 & \bf \small BLEU & \bf \small ROUGE-1 & \bf \small ROUGE-2 & \bf \small ROUGE-L  \\
\hline \hline
(a) & Baseline: Iterative training & 71.14 & 55.05 & 55.37 & 27.95 & 39.90 \\
(b) & Dual supervised learning, $\lambda = 0.1$ & \bf 72.32 & \bf 57.16 & \bf 56.37 & \bf 29.19 & \bf 40.44 \\
(c) & Dual supervised learning, $\lambda = 0.01$ & 72.08 & 55.07 & 55.56 & 28.42 & 40.04 \\
(d) & Dual supervised learning, $\lambda = 0.001$ & 71.71 & 56.17 & 55.90 & 28.44 & 40.08 \\
(e) & Dual supervised learning w/o MADE  & 70.97 & 55.96 & 55.99 & 28.74 & 39.98 \\
\hline
  \end{tabular}
\vspace{-1mm}
\caption{The NLU performance reported on micro-F1 and the NLG performance reported on BLEU, ROUGE-1, ROUGE-2, and ROUGE-L of models (\%).}
\vspace{-3mm}
\label{tab:results}
\end{table*}

\begin{comment}
\subsection{Optimization}
With the estimation of semantics $\hat{P}(x)$ and natural language $\hat{P}(y)$, the multi-objective problem is optimized in a supervised fashion.
\end{comment}

\section{Experiments}
To evaluate the effectiveness of the proposed framework, we conduct the experiments, the settings and analysis of the results are described as follows.

\subsection{Settings}
The experiments are conducted in the benchmark E2E NLG challenge dataset~\cite{novikova2017e2e}, which is a crowd-sourced dataset of 50k instances in the restaurant domain.
Our models are trained on the official training set and verified on the official testing set. 
Each instance is a pair of a semantic frame containing specific slots and corresponding values and an associated natural language utterance with the given semantics.
% For example, a semantic frame with the slot-value pairs ``\texttt{name[Bibimbap House], food[English], priceRange[moderate], area [riverside], near [Clare Hall]}''  corresponds to the target sentence ``\emph{Bibimbap House is a moderately priced restaurant who's main cuisine is English food. You will find this local gem near Clare Hall in the Riverside area.}''. 
The data preprocessing includes trimming punctuation marks, lemmatization, and turning all words into lowercase. 

Although the original dataset is for NLG, of which the goal is to generate sentences based on the given slot-value pairs, we further formulate a NLU task as predicting slot-value pairs based on the utterances, which is a multi-label classification problem.
Each possible slot-value pair is treated as an individual label, and the total number of labels is 79.
To evaluate the quality of the generated sequences regarding both precision and recall, for NLG, the evaluation metrics include BLEU and ROUGE (1, 2, L) scores with multiple references, while F1 score is measured for the NLU results.

% model settings 
\subsection{Model Details}
The model architectures for NLG and NLU are a gated recurrent unit (GRU)~\cite{cho2014learning} with two identical fully-connected layers at the two ends of GRU.
Thus the model is symmetrical and may have semantic frame representation as initial and final hidden states and sentences as the sequential input.

In all experiments, we use mini-batch \textit{Adam} as the optimizer with each batch of 64 examples, 10 training epochs were performed without early stop, the hidden size of network layers is 200, and word embedding is of size 50 and trained in an end-to-end fashion.

\subsection{Results and Analysis}
The experimental results are shown in Table \ref{tab:results}, where each reported number is averaged over three runs.
%results on the official testing set from three different models. 
The row (a) is the baseline that trains NLU and NLG separately and independently, and the rows (b)-(d) are the results from the proposed approach with different Lagrange parameters.

The proposed approach incorporates probability duality into the objective as the regularization term.
To examine its effectiveness, we control the intensity of regularization by adjusting the Lagrange parameters.
The results (rows (b)-(d)) show that the proposed method outperforms the baseline on all automatic evaluation metrics.
Furthermore, the performance improves more with stronger regularization (row (b)), demonstrating the importance of leveraging duality.

In this paper, we design the methods for estimating marginal distribution for data in NLG and NLU tasks: language modeling is utilized for sequential data (natural language utterances), while the masked autoencoder is conducted for flat representation (semantic frames).
The proposed method for estimating the distribution of semantic frames considers complex and implicit dependencies between semantics by ensemble of multiple decomposition of joint distribution.
In our experiments, the empirical marginal distribution is the average over the results from 10 different masks and orders; in other words, 10 types of dependencies are modeled.
The row (e) can be viewed as the ablation test, where the marginal distribution of semantic frames is estimated by considering slot-value pairs independent to others and statistically computed from the training set.
The performance is worse than the ones that model the dependencies, demonstrating the importance of considering the nature of input data and modeling data distribution via the masked autoencoder.

We further analyze understanding and generation results compared with the baseline model.
In some cases, it is found that our NLU model can extract the semantics of utterances better and our NLU model can generate sentences with richer information based on the proposed learning scheme.
% In some cases, it is found that our NLU model extracts more slots in order to cover all salient slots for better generation.
% On the other hand, our NLG model tends to include all slot values for aligning with the semantics NLU extracts.
In sum, the proposed approach is capable of improving the performance of both NLU and NLG in the benchmark data, where the exploitation of duality and the way of estimating distribution are demonstrated to be important.

\section{Conclusion}
This paper proposes a novel training framework for natural language understanding and generation based on dual supervised learning, which first exploits the duality between NLU and NLG and introduces it into the learning objective as the regularization term.
Moreover, expert knowledge is incorporated to design suitable approaches for estimating data distribution.
The proposed methods demonstrate effectiveness by boosting the performance of both tasks simultaneously in the benchmark experiments.

\section*{Acknowledgements}
We thank the anonymous reviewers for their insightful feedback on this work. 
This work was financially supported from the Young Scholar Fellowship Program by Ministry of Science and Technology (MOST) in Taiwan, under Grant 108-2636-E-002-003 and 108-2634-F-002-019.

\bibliography{acl2019}
\bibliographystyle{acl_natbib}
\end{document}